\documentclass[letterpaper, 10 pt, conference]{ieeeconf} 

\IEEEoverridecommandlockouts                             

\usepackage{algorithm}
\usepackage{algorithmic}
\usepackage{amsmath}
\usepackage{amsfonts}
\usepackage{newfloat}
\usepackage{listings}
\usepackage{color}
\usepackage{float}
\usepackage{multirow}

\usepackage{graphics} 
\usepackage{epsfig} 
\usepackage{mathptmx} 
\usepackage{times} 
\usepackage{amsmath} 
\usepackage{hyperref}

\title{\LARGE \bf Subequivariant Reinforcement Learning Framework for Coordinated Motion Control}

\author{Haoyu Wang$^{1\dagger}$, Xiaoyu Tan$^{2\dagger}$, Xihe Qiu$^{1\dagger*}$ and Chao Qu$^{2*}$
\thanks{$\dagger$ Both authors contributed equally to this work}
\thanks{* Corresponding author: Xihe Qiu, Chao Qu, qiuxihe@sues.edu.cn, quchao\_tequila@inftech.ai}
\thanks{$^{1}$Haoyu Wang and Xihe Qiu are with Shanghai University of Engineering Science, Shanghai, China
      }  
\thanks{$^{2}$Xiaoyu Tan and Chao Qu are with INF Technology (Shanghai) Co., Ltd
       } 
}

\begin{document}

\maketitle
\thispagestyle{empty}
\pagestyle{empty}

\begin{abstract}

Effective coordination is crucial for motion control with reinforcement learning, especially as the complexity of agents and their motions increases. However, many existing methods struggle to account for the intricate dependencies between joints. We introduce CoordiGraph, a novel architecture that leverages subequivariant principles from physics to enhance coordination of motion control with reinforcement learning. This method embeds the principles of equivariance as inherent patterns in the learning process under gravity influence, which aids in modeling the nuanced relationships between joints vital for motion control. Through extensive experimentation with sophisticated agents in diverse environments, we highlight the merits of our approach. Compared to current leading methods, CoordiGraph notably enhances generalization and sample efficiency.

\end{abstract}


\section{INTRODUCTION}
Reinforcement learning (RL) is a prominent approach for enabling intelligent agents to acquire skills for intricate tasks via iterative trial and error\cite{heuillet2021explainability}. However, managing the coordinated movements of multiple joints through RL, especially in agents navigating complex physical environments like multi-joint robotic systems\cite{korber2021comparing}, is challenging\cite{chen2022reinforcement,liu2020distance}. Traditional RL techniques\cite{ladosz2022exploration} often address this problem through the lens of the curse of dimensionality and training instabilities \cite{qiu2022latent}. Nevertheless, these techniques often overlook the interactions between joints and the physical principles prevalent in most agent operating conditions.

Graph neural networks (GNNs) have demonstrated potential in RL for coordinated motion control by representing node internal interactions\cite{park2021learning}, \cite{chen2022model}. However, they do have some challenges prevent the GNN in practical utilization\cite{chai2020motor}. Specifically, GNNs can occasionally find it difficult to recognize dynamic symmetries and maintain equivariance in joint interactions, which can result suboptimal coordination performance\cite{wang2018nervenet}. They also often face challenges in exploration, limiting their efficiency in discovering rewards in novel scenarios\cite{wu2021ironman,dewitt2020deep}. Additionally, GNNs typically require significant training data and time to reach the desired performance\cite{hart2020graph}, which can constrain their practical adaptability\cite{shan2021reinforcement}.

To intergrete prior knowledge of symmetry-related prior in the RL-based motion control, many equivariant techniques have been proposed\cite{blake2021snowflake,zhao2021physics}. However, most of equivariant neural networks are designed under the assumption of specific symmetries in the input graph data. This design choice can restrict their utility, especially for graph structures that exhibit global symmetries\cite{castillo2020hybrid}. In complex graph datasets, these methods might not accurately capture intricate symmetry patterns, potentially hindering performance and generalization\cite{wang2023scientific,wang2023neural}. 

\begin{figure}[h]
    \centering
    \begin{tabular}{@{\extracolsep{\fill}}c@{}c@{}c@{}c@{\extracolsep{\fill}}}
            \includegraphics[width=0.25\linewidth, height=3cm]{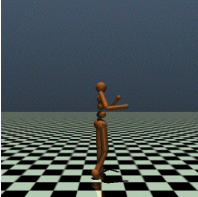} &
            \includegraphics[width=0.25\linewidth, height=3cm]{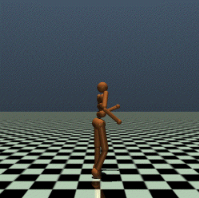} &
            \includegraphics[width=0.25\linewidth, height=3cm]{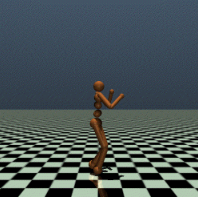} &
            \includegraphics[width=0.25\linewidth, height=3.018cm]{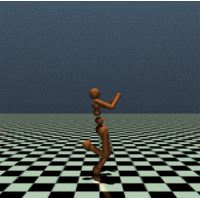}\\
    \end{tabular}
    \caption{Training a humanoid agent in the MuJoCo environment with the objective of enabling it to transition from an inability to stand to coordinated joint movements.}
    \label{p1}
    \vspace{-2ex}
\end{figure}

In this study, we propose \textbf{CoordiGraph}, a novel approach that leverages coordinated subequivariant networks for joint motion control in reinforcement learning. Our method addresses the limitations of GNNs and equivariant technique in coordinating motion control with RL. By incorporating the concept of subequivariance into the GNNs framework, CoordiGraph effectively models symmetries and subequivariance within agent joints. This enhances agent performance and efficiency in cooperative motion learning tasks, as illustrated in Figure \ref{p1}. Our model ensures the preservation of input data symmetry and accurately captures subequivariance properties between joints, enabling agent actions to maintain symmetry and equivariance.

Through extensive experiments conducted on diverse reinforcement learning control benchmark tasks, we empirically demonstrate the effectiveness in coordinating agent motion, showcasing the significant benefits of incorporating subequivariant principles with reinforcement learning in motion control tasks by achieving improved coordination and learning efficiency.

\section{Related Work}
\subsection{Reinforcement Learning on Motion Control}
Reinforcement learning is a method of learning optimal strategies by interacting with the environment, using reward signals to guide the actions of an agent and maximize cumulative rewards\cite{levine2020offline}. It has applications in domains like robot control and game AI\cite{zhao2020sim}. The goal is to find the best policy that maximizes the agent's cumulative rewards. Various algorithms have been proposed in the field of reinforcement learning for motion control\cite{moerland2023model}, including deep reinforcement learning methods like Deep Q-Network for training agents in Atari games\cite{moreno2019performing}. Evolution Strategies have been used to train agents for complex behaviors like walking and running\cite{grillitsch2020trinity,li2021reinforcement}. Transfer learning from simulated to real environments has also been achieved. Genetic algorithms and particle swarm optimization\cite{lakshmanan2020complete,li2021constrained} have been used to optimize coordinated movements among different components of agents\cite{wang2019pso,zhou2022maintenance}. Hierarchical approaches, where higher-level policies guide lower-level policies\cite{schilling2019approach}, have been successful in tasks requiring hierarchical organization, such as multi-agent navigation and cooperative transportation\cite{fu2019deep,li2020deep}.

\subsection{Coordinating Joint Movements for Graph}
Graph neural networks (GNN) are widely used in reinforcement learning to model interactions between entities in a graph structure\cite{cai2021jolo}. GNN leverage information propagation through edges and nodes to capture precise dependencies between agents in coordinated tasks\cite{tu2022joint,ding2022towards}. Variants such as graph convolutional network aggregate information from neighboring agents, enabling localization and recognition of temporal actions\cite{zhang2022graph}. Graph attention networks assign weights to features based on relevance, accurately predicting future trajectories by representing positions and relationships as a graph\cite{shao2021graph}. GraphSAGE uses sampling strategies to process information from fixed-sized agent neighborhoods\cite{dai2022cooperative}, achieving better generalization capabilities regardless of graph size and structure\cite{jiang2018graph,gammelli2021graph}.

\subsection{Advancements in Subequivariant Techniques}
Subequivariant neural networks is an extension of neural Networks that handles symmetry and subequivariance\cite{han2022learning}. Subequivariant network enhances the learning capability of traditional neural Networks by introducing the concept of subequivariance. It builds upon the idea of equivariant graph neural networks (EGNN) that handle graph data with symmetry\cite{satorras2021e}. EGNN achieves this by designing graph convolution operations with symmetry and preserving the symmetry properties of the input graph during learning. Researchers have also explored other improvement measures, such as graph matching networks (GMN) that handle local symmetry by applying equivariance operations like rotation and translation\cite{huang2022equivariant}. Leveraging subgraph information and incorporating equivariance into cross-domain graph neural networks has shown effective generalization to unseen target domains\cite{yu2022finding}. Other studies have also investigated reinforced learning\cite{wang2022equivariant} and low-dimensional feature extraction\cite{fu2022reinforced} in the context of equivariant graph neural networks\cite{dym2022low}.

\section{Methodology}
We define the state space \(S\) as an \(n\)-dimensional vector representing the state of each joint: \(S = [s_1, s_2, \ldots, s_n]\). 

\textbf{Designing between Agents and Environment} The agent, modeled as a discrete graph structure, can perform actions within a certain range, such as applying force or torque \cite{arulkumaran2017deep}. The action space \(A\) is an \(n\)-dimensional vector representing the actions of each joint: \(A = [a_1, a_2, \ldots, a_n]\). By designating the body node as the coordinate system, the joint nodes represent the degrees of freedom between them. For example, Walker2d-v2, the root node determines the agent's position in MuJoCo.

The policy function \(\pi\) maps the state space \(S\) to a probability distribution over the action space \(A\). \(\pi(a|s)\) represents the probability of selecting action \(a\) given state \(s\). The value function \(V\) estimates the expected return under state \(s\), denoted as \(V(s)\) mapping the state space \(S\) to the state value.

\begin{figure}[h]
    \centering
    \begin{tabular}{@{\extracolsep{\fill}}c@{}c@{\extracolsep{\fill}}}
            \includegraphics[height=0.48\linewidth,width=0.48\linewidth]{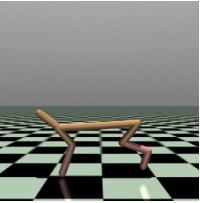} &
            \includegraphics[height=0.48\linewidth,width=0.5\linewidth]{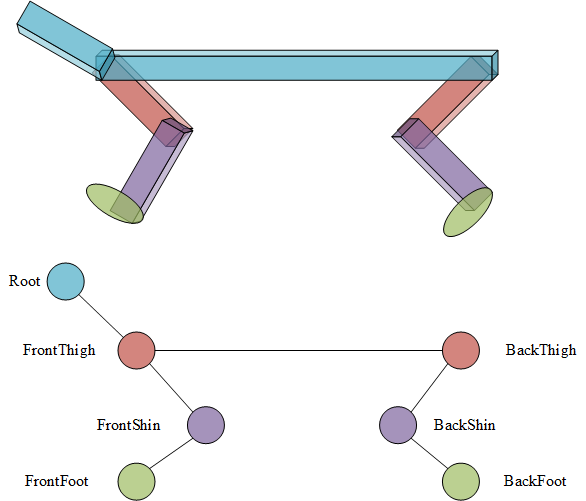}\\
            (a)Mujoco & (b)Format\\
    \end{tabular}
    \caption{Modeling the environment in MuJoCo, where agents possess multiple hierarchical joints. Simple graph neural networks are insufficient to fully capture the interaction features among joints. Introducing subequivariance requires hierarchical classification for different joints, as depicted in this figure.}
    \label{f2}
    \vspace{-2ex}
 \end{figure}

\textbf{Subequivariant Learning Networks} We design our model to explore equivariant properties by incorporating external fields, such as gravity, in a reasonable manner. It decomposes the graph structure into subgraphs based on equivariant properties and propagates equivariant information between these subgraphs to handle graph feature data across different joints of an intelligent agent, as shown in Figure \ref{f2}. 
\begin{equation}
 h_i^{(l+1)} = \sigma\left(\sum_{j \in N(i)} f(h_i^{(l)}, h_j^{(l)}, E_{ij})\right),
 \label{e1}
\end{equation} here $h_i^{(l)}$ represents the feature representation of the central node $i$ in the $l$-th layer, $N(i)$ denotes the neighboring nodes of the central node $i$, $f$ is the aggregation function, and $E_{ij}$ represents the external field information between the central node $i$ and the neighboring node $j$.

CoordiGraph introduces an Object-aware Message Passing mechanism for learning physical interactions between objects of different shapes. This mechanism handles object properties, sizes, and shapes, enabling hierarchical modeling and improving the model's ability to handle complex interactions. 
\begin{equation}
m^{(l+1)}_{ij} = g(h^{(l)}_i, h^{(l)}_j, s^{(l)}_i, s^{(l)}_j),
\label{e2}
\end{equation} $m_{ij}^{(l+1)}$ represents the object-aware message between the central node $i$ and the neighboring node $j$ in the layer of $(l+1)$. $s_i^{(l)}$ and $s_j^{(l)}$ represent the shape representations of the central node $i$ and the neighboring node $j$ in the $l$-th layer. The function $g$ integrates node features and shape information, evolving with subsequent policy updates. Its purpose is to capture the physical interactions between objects of different shapes by combining their features and shape information.
\begin{equation}
\begin{aligned}
 m_v^{(l)} &= \sum_{u \in N(v)} \frac{1}{\sqrt{|N(v)||N(u)|}} \cdot W^{(l)} \cdot h_u^{(l-1)} \\
 h_v^{(l)} &= \sigma(m_v^{(l)} + W_0^{(l)} \cdot h_v^{(l-1)}) \\
\end{aligned}
\end{equation}

It calculates the aggregated message $m_v^{(l)}$ for node $v$ in the $l$-th layer by summing the weighted hidden states $h_u^{(l-1)}$ of its neighboring nodes $u$. The weights are normalized by the square root of the product of the degrees of the nodes, ensuring a balanced influence from different degrees. The weights are applied through the weight matrix $W^{(l)}$. The update rule for the hidden state $h_v^{(l)}$ of node $v$ in the $l$-th layer combines the aggregated message $m_v^{(l)}$ with the previous hidden state $h_v^{(l-1)}$ using the self-loop weight matrix $W_0^{(l)}$.

In the subequivariant neural network model, we introduce two time-related concepts. Firstly, there is the classical notion of time, represented by the time step $t$ in the environment. Additionally, we define the internal propagation step as $\tau$, which represents the series of steps performed by the model to determine node actions based on environmental observations within each time step.
\begin{equation}
m_{uv}^{(t)} = f(h_{tu}, h_{tv}),
\label{e3}
\end{equation} here $m_{uv}^{(t)}$ represents the message vector from node $u$ to node $v$, $h_{tu}$ and $h_{tv}$ denote the state vectors of node $u$ and node $v$ respectively in propagation step $t$. 
\vspace{1ex}

During time step $t$, each node $u$ performs information propagation by computing message vectors with its neighboring node $v$. By utilizing the state vectors $h_{tu}$ and $h_{tv}$ of nodes $u$ and $v$ respectively, node $u$ calculates the message vector $m_{uv}^{(t)}$ using the function $f$. These message vectors update the node's state and facilitate further information propagation in subsequent propagation steps.
\begin{equation}
h_i^{(l+1)} = g(h_i^{(l)}, h_i^{(l-1)}),
\end{equation} in this equation $h_i^{(l+1)}$ represents the updated feature representation of node $i$ in the next layer. 

The function $g$ combines the current feature representation $h_i^{(l)}$ with the previous feature representation $h_i^{(l-1)}$ to generate the updated representation.

To avoid the loss of interaction features and enhance coordination capabilities between agent joints, we propose a novel feature representation approach. This approach preserves the properties of vector feature vectors while stacking them together to retain the interaction information. 
\begin{equation}
{u}_i = {v}_i - \frac{1}{N}\sum_{j=1}^{N}{v}_j,
\end{equation} ${v}_i$ represents the vector feature vector with index $i$. We convert each vector feature vector into a translation-invariant vector ${u}_i$ by subtracting the average of all feature vectors $\frac{1}{N}\sum_{j=1}^{N}{v}_j$ from ${v}_i$. Here, $N$ denotes the number of feature vectors.
\vspace{1ex}

\textbf{Reinforcement Algorithms and Optimization} We employ the proximal policy optimization to update the policy function parameters for better adaptation to the environment. The advantage function $A$, which measures the relative advantage of each action, is defined as the difference between the action-value function $Q(s, a)$ and the state-value function $V(s)$.
\begin{equation}
A(s, a) = Q(s, a) - V(s)
\end{equation}

The objective is to maximize the expected cumulative reward by adjusting the policy parameters $\theta$. 
\begin{equation}
R(\pi_{\theta}) = \sum_{t=0}^T \alpha^t r(s_t, a_t),
\label{e7}
\end{equation} $\alpha$ represents the discount factor, $t$ denotes the number of time steps. 

At each time step $t$, the agent selects an action $a_t$ based on the current state $s_t$ and receives a reward $r(s_t, a_t)$ from the environment. The environment transitions to the next state according to the transition probability $P(s_{t+1}|s_t)$.
\begin{equation}
  P(a^\tau | s^\tau) = \prod_{u \in \mathcal{O}} P_u(a_u^\tau | s^\tau) 
\label{e10}
\end{equation}
\begin{equation}
P_u(a_u^\tau | s^\tau) = \frac{1}{\sqrt{2 \pi \sigma_u^2}} \exp\left(\frac{(a_u^\tau - \mu_u)^2}{2 \sigma_u^2}\right)
\label{e11}
\end{equation}

The agent's objective is to learn policy parameters $\theta$ that maximize the cumulative reward during interaction with the environment. This involves continuously updating the policy parameters through interactions to improve decision-making and maximize long-term cumulative reward \cite{le2022deep}.
\vspace{1ex}

The proximal policy optimization algorithm aims to maximize the advantage function $A$ while constraining policy changes to ensure stability. The objective function of PPO is defined as:
\begin{equation}
\theta_{\text{new}} = \arg\max_{\theta} {E}{\pi{\theta_{\text{old}}}} \left[ \frac{\pi_{\theta}(a|s)}{\pi_{\theta_{\text{old}}}} \cdot A^{\pi_{\theta_{\text{old}}}}(s, a) \right], 
\end{equation} $\theta_{\text{new}}$ represents the updated policy parameters, and $\theta_{\text{old}}$ represents the old policy parameters. $\pi_{\theta}(a|s)$ is the policy that selects action $a$ given state $s$, while $\pi_{\theta_{\text{old}}}$ represents the policy based on the old parameters. 
\vspace{1ex}

$A^{\pi_{\theta_{\text{old}}}}(s, a)$ is the advantage function based on the old policy, which estimates the advantage of selecting action $a$ in state $s$ relative to the old policy. The objective function maximizes the expected value of the ratio between the new and old policies, weighted by the advantage function.
\begin{equation}
\begin{aligned}
\tilde{J}(\theta) &= \mathbb{E}_{\pi_\theta}\left[\sum_{\tau=0}^{\infty} \min\left(\hat{A}^\tau r^\tau(\theta), \hat{A}^\tau \operatorname{clip}\left(r^\tau(\theta), 1-\epsilon, 1+\epsilon\right)\right)\right] \\
&\quad - \beta \mathbb{E}_{\pi_\theta}\left[\sum_{\tau=0}^{\infty} \operatorname{KL}\left[\pi_\theta(\cdot|s^\tau) \parallel \pi_{\theta_{\text{old}}}(\cdot|s^\tau)\right]\right] \\
&\quad - \alpha \mathbb{E}_{\pi_\theta}\left[\sum_{\tau=0}^{\infty} \left(V_\theta(s^\tau) - V(s^\tau)^{\text{target}}\right)^2\right]
\end{aligned}
\end{equation}

The objective of policy updates in PPO is to maximize the expected ratio of action probabilities, weighted by the advantage under the old policy. This approach ensures that the new policy gradually improves its performance while staying close to the old policy, represented by $\pi_{\theta_{\text{old}}}$ \cite{nian2020review}. 

\begin{algorithm}[h]
\caption{Total Algorithm}\label{alg:HGAJ}
\begin{algorithmic}
		\STATE {\textbf{Inputs}:Input the state of each joint $S$, the actions of each joint $A$ and the policy function \(\pi\).\\
        \STATE {\textbf{Require}:Update $S$ and $A$, extract the feature representation of the node $h_i^{(l)}$, object-aware message $m_{ij}^{(l+1)}$ between the central node and the neighboring node, updated policy parameters $\theta_{\text{new}}$. }\\
        \STATE{\textbf{Function Designing Agents and Environment($V$, $A$, $S$)}}\\
        \STATE{ $\pi \leftarrow S, A$}, maping the policy function $\pi$ a probability distribution and representing the probability of selecting action.}\\
        \STATE{ $V(s) \leftarrow S$}, estimating the expected return under state.\\
         \STATE{ $s, r \leftarrow \pi, S, A$}, agent selects an action from the action space based on the current state and interacts with the environment by executing action until finding a policy that maximizes the expected reward.\\
        \textbf{Function Subequivariant Learning($h_i^{(l)}$, $h_j^{(l)}$, $t$)}\\
        \STATE{ $E_{ij} \leftarrow h_i,h_j$, representing the external field information between the central node $i$ and the neighboring node $j$}.\\
        \STATE{ $m^{(l+1)}_{ij} \leftarrow h_i^{(l)}, h_j^{(l)}$, using the equation \ref{e2}, representing the object-aware message between the central node $i$ and the neighboring node $j$ in the layer indexed by $(l+1)$.}\\
        \STATE{ {$m_{uv}^{(t)} \leftarrow m^{(l+1)}_{ij}, t, f$}, updating the
        node’s state and facilitate further information propagation in subsequent propagation steps through equation \ref{e3}.}\\
        \textbf{Function Training and Optimization ($s_t$, $a_t$, $\theta$, $\pi_{\theta}(a|s))$}\\
        \STATE{ $\theta_{\text{new}} \leftarrow {\theta_{\text{old}}}, \pi_{\theta}(a|s))$, the advantage function is used to estimate the advantage of selecting action $a_t$ in state $s_t$ relative to the old policy through equation \ref{e7}.}\\
\end{algorithmic}
\end{algorithm}

\section{Experiments and Simulations results}

In this section, we present the results of our experiments evaluating the effectiveness of the subequivariant-based neural network in improving agent motion in cooperative reinforcement learning tasks. We compare its performance with traditional graph networks and emphasize the advantages of integrating second-order invariance into the network architecture. These experiments cover various benchmark task environments to evaluate the generalizability and robustness of our model.

\subsection{Experimental Setup}
We selected multiple agents from MuJoCo, including Hopper-v2, Humanoid-v2\cite{todorov2012mujoco}, HalfCheetah-v2, Centipede-v1\cite{wang2018nervenet}, and Walker2d-v2. The goal was to apply reinforcement learning to coordinate the joint movements of these agents. Tasks such as standing, coordinated joint manipulation, and normal locomotion were designed to test different aspects of coordination.

\textbf{Traditional Graph} in our experiments uses a network architecture of [512, 512], a learning rate of 3e-4, a hidden state size of 256, separate output networks tailored to each agent's coordination requirements, and 6 propagation steps. 


\subsection{Parameter Architecture and Evaluation Metrics}
Experiments used an NVIDIA GeForce RTX 3080 GPU and an 8-core CPU. Each validation training session lasted around 16 hours. To ensure fair comparison, we maintained consistent batch size, training iterations, and performance metrics. The table \ref{t1} displays the hyperparameter settings for our modules.
\begin{table}[h]
\caption{Parameter architecture and evaluation metrics}
\resizebox{\linewidth}{!}{
\begin{tabular}{l|c}
\hline 
CoordiGraph & Value Tried \\
\hline 
Gradient clipping &  0.05, 0.1, 0.2 \\
Network Shape & {$[128,128], [256,256], [512,512]$} \\
Learning rate & 1e-4, 3e-4 \\
Hidden state size & 128, 256\\
Size of Nodes' Hidden size & $32,64,128$ \\
Output Network & Shared, Separate \\
Add Skip-connection from / to root & Yes, No  \\
Number of Propogation Steps & $4,5,6$ \\
Matrix embedding size & 32×32, 64×64 \\
Learning rate scheduler & adaptive, constant \\
\hline
\end{tabular}
}
\label{t1}
\end{table}

\subsection{Main Results}
We adopt a reinforcement learning framework to train agents in a coordination task. By interacting with the environment, the agents receive rewards based on their actions and update their policies using policy gradient methods. We employ a variant of proximal policy optimization as the training algorithm, which has been proven effective in multi-agent reinforcement learning \cite{carroll2019utility}.

\begin{figure*}[thbp!]
    \centering
    \begin{tabular}{@{\extracolsep{\fill}}c@{}c@{\extracolsep{\fill}}}
            \includegraphics[width=0.48\linewidth, height=3.5cm]{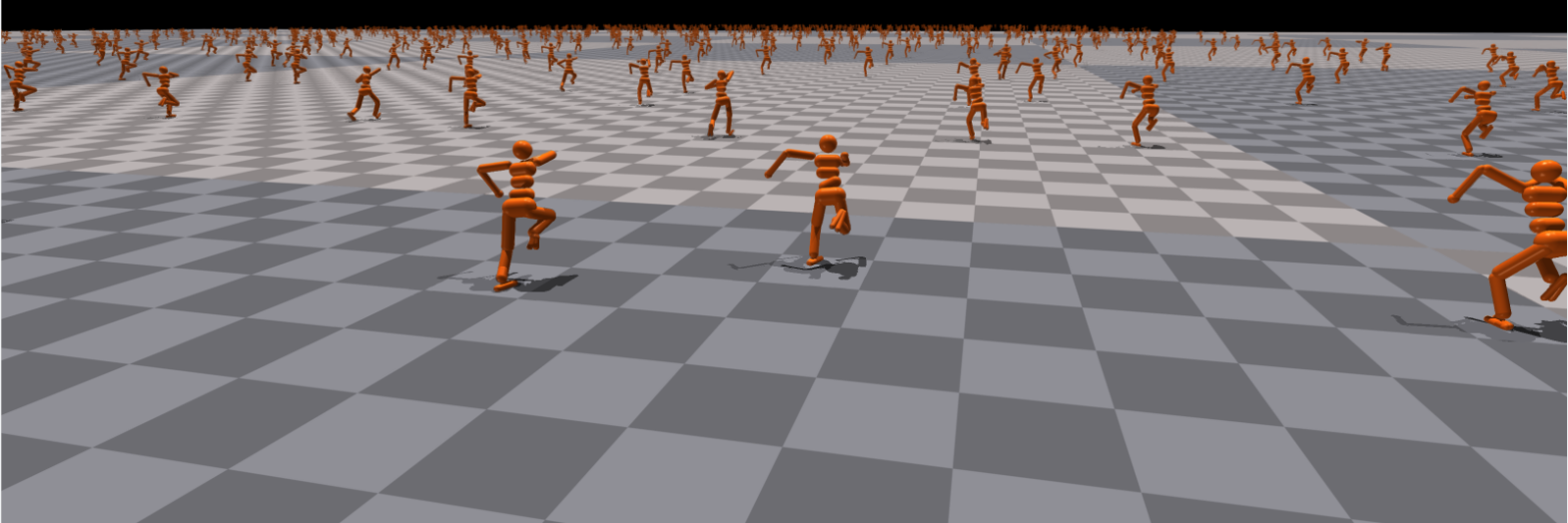} &
            \includegraphics[width=0.48\linewidth, height=3.5cm]{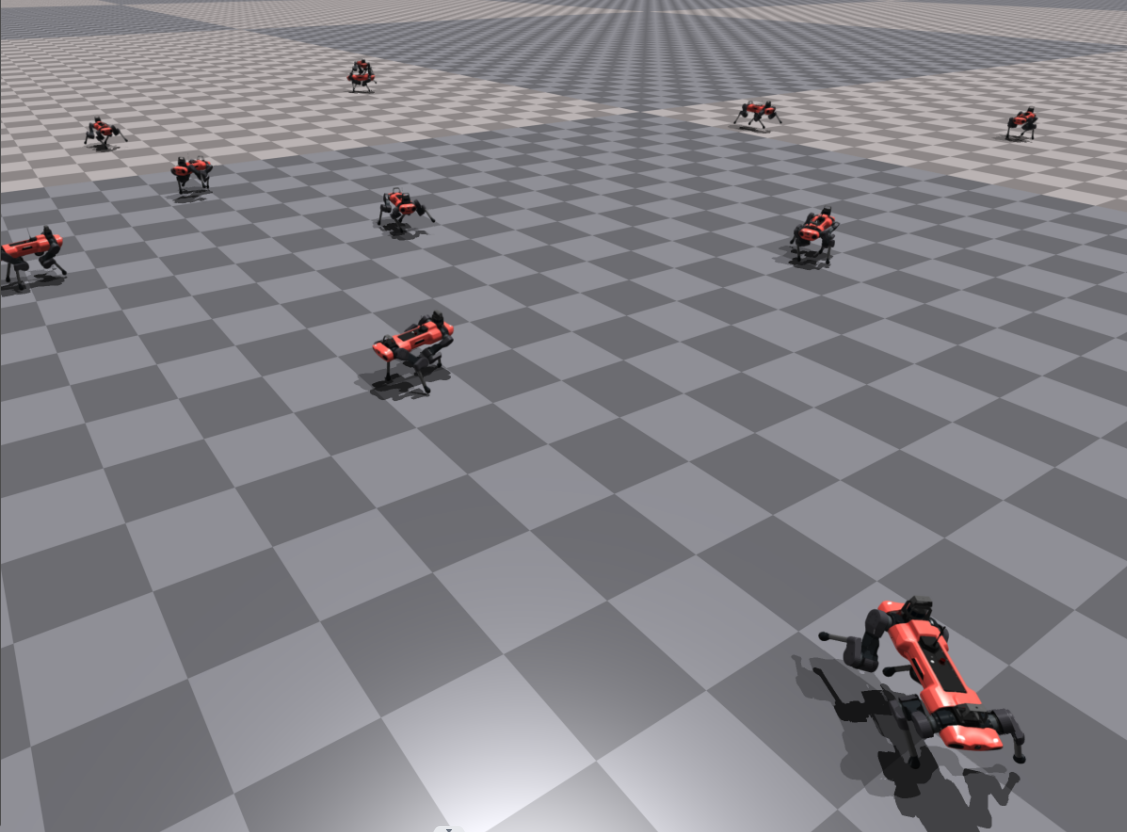}\\
    \end{tabular}
    \caption{We conducted large-scale training in a simulated environment, incorporating various environments and agents to ensure the generalizability and practicality of the model's performance.}
    \label{f3}
    \vspace{-3ex}
\end{figure*}

Our experimental results demonstrate that our method outperforms existing techniques in multi-joint robot control as shown in Table \ref{t2}. We observe significant improvements in coordination accuracy, exploration capability, and reward acquisition. These findings highlight the effectiveness of our approach in addressing the challenges of the Mujoco multi-joint robot control problem in Figure \ref{f3}.

We conducted experiments to compare the performance of CoordiGraph and traditional graph networks in motion control tasks using reinforcement learning. The results, as depicted in the figure, consistently demonstrate the superior performance of CoordiGraph over traditional graph networks across all coordination tasks, with a notable advantage in terms of reward acquisition.

The superior performance of CoordiGraph in coordination tasks can be attributed to its incorporation of second-order variance characteristics. By effectively capturing dynamic variations in symmetry and isotropy within the joints of intelligent agents, CoordiGraph enhances coordination accuracy. In contrast, traditional graph networks, such as GNN, exhibit weaker performance in this aspect.

Furthermore, CoordiGraph showcases superior exploration and coordination capabilities, enabling it to adapt and learn more effectively in unfamiliar scenarios. In contrast, traditional graph network demonstrates weaker abilities in exploration and coordination. The advantages of CoordiGraph in motion control tasks extend beyond reward acquisition. It excels in accurately predicting actions and joint angle positions of intelligent agents during task execution, facilitating precise coordination. This heightened accuracy empowers CoordiGraph to adapt more efficiently to complex environments and task requirements, ultimately enhancing task efficiency.
\begin{table*}[h]
\caption{Results of our graph network's efficiency}
\resizebox{0.95\textwidth}{!}{
\begin{tabular}{c|c|c|c|c|c|c|c}
\hline 
Model & \multicolumn{7}{c}{Avg\_Reward} \\
\hline 
Environment & Centipede-Four & Humanoid & HalfCheetah & Centipede-Six & Walker2D & Hopper & Centipede-Ten \\
\hline 
\multirow{5}{*}{CoordiGraph} 
& 5792.07 & 1088.36 & 7803.52 & 6325.47 & 4376.90 & 3693.25 & 4395.71 \\
& 5960.48 & 998.05 & 7612.81 & \textbf{6732.12} & 4685.21 & 3323.49 & 4632.08 \\
& 5868.34 & \textbf{1110.37} & \textbf{7937.41} & 6538.09 & 4615.75 & 3594.31 & \textbf{4801.35} \\
& \textbf{6042.98} & 1025.48 & 7452.94 & 6694.68 & \textbf{4953.14} & \textbf{3774.04} & 4438.92 \\
& 5586.42 & 1101.91 & 7567.58 & 6321.57 & 4831.48 & 3608.21 & 4735.86 \\
\hline 
\multirow{4}{*}{Traditional Graph} 
& 3927.33 & 938.62 & 5364.29 & 4621.44 & 3683.17 & 2367.56 & 3808.19 \\
& 3643.29 & 967.53 & 5646.37 & 5012.19 & 3701.21 & 2965.28 & 3329.85 \\
& 4256.83 & 878.15 & 5491.42 & 4832.53 & 3358.91 & 2838.71 & 3894.24 \\
& 3847.51 & 908.89 & 4992.81 & 5177.47 & 3401.39 & 2547.19 & 3674.52 \\
& 4168.17 & 899.76 & 5513.24 & 4937.29 & 3786.58 & 2636.42 & 3752.39 \\
\hline
\end{tabular}
}
\label{t2}
\vspace{-3ex}
\end{table*}

In the subsequent phase, we use ablation experiments to demonstrate the superiority of CoordiGraph. By surpassing existing neural networks in coordinating motion, our model validates its effectiveness in motion control.

\textbf{Directionality of Intelligent Agent Motion}
In our experiments, we trained two CoordiGraph models under different environmental conditions: one with directionality constraints and one without. Directionality constraints limited the model's learning and exploration to specific directions, while the unconstrained model could learn and explore in all directions.

\begin{figure}[thbp!]
\raggedright
    \begin{minipage}[t]{0.6\linewidth}
        \raggedright
        \includegraphics[width=1.7\linewidth]{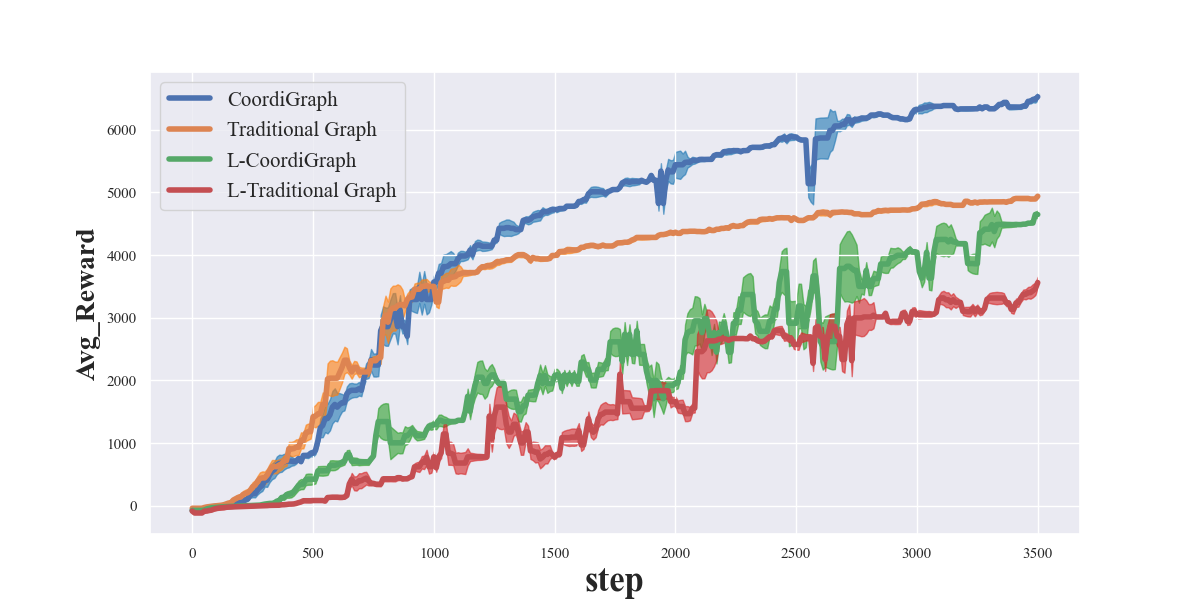}
    \end{minipage}
    \begin{minipage}[t]{0.6\linewidth}
\raggedright
      \raggedright
        \includegraphics[width=1.7\linewidth]{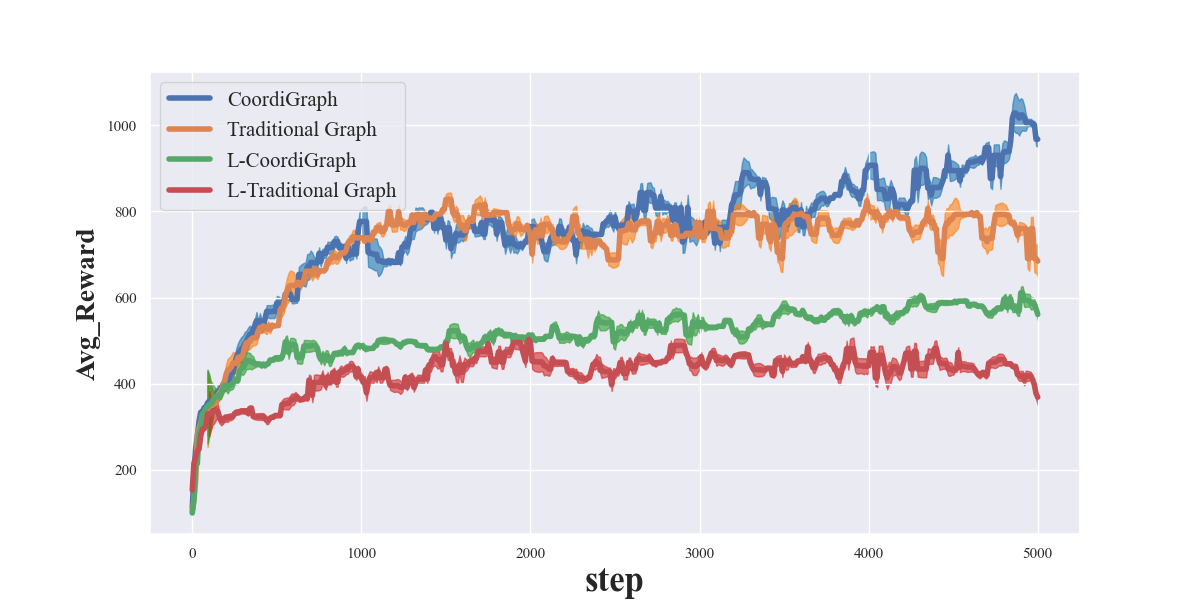}
    \end{minipage}
    \caption{Results of the dynamics of intelligent agent motion}
    \label{Ab1}
    \vspace{-1ex}
 \end{figure}

We tested both models in the same environment and compared their performance under different directionality constraint conditions. By observing the model's behavior, we evaluated the adaptability and generalization abilities of the CentipedeSix and Humanoid environments in Figure \ref{Ab1}, providing insights into the impact of directionality constraints on learning and performance.

The results show that the model trained with directionality constraints outperforms in testing. By focusing on specific behaviors and optimizing them through trial and error, the model gains a better understanding of the environment and takes actions that lead to higher rewards.

In contrast, the model trained without directionality constraints faces challenges in optimizing its behavior due to increased uncertainty and randomness during training. It may struggle to achieve the same high rewards as when directionality is controlled.

Imposing directionality constraints allows the model to concentrate on learning and optimizing specific behaviors, leading to improved learning outcomes during testing. This constraint enhances the model's understanding of the environment and enables it to take appropriate actions to maximize rewards.
\begin{figure}[h]
\raggedright
    \includegraphics[width=0.5\textwidth,height=4.8cm]{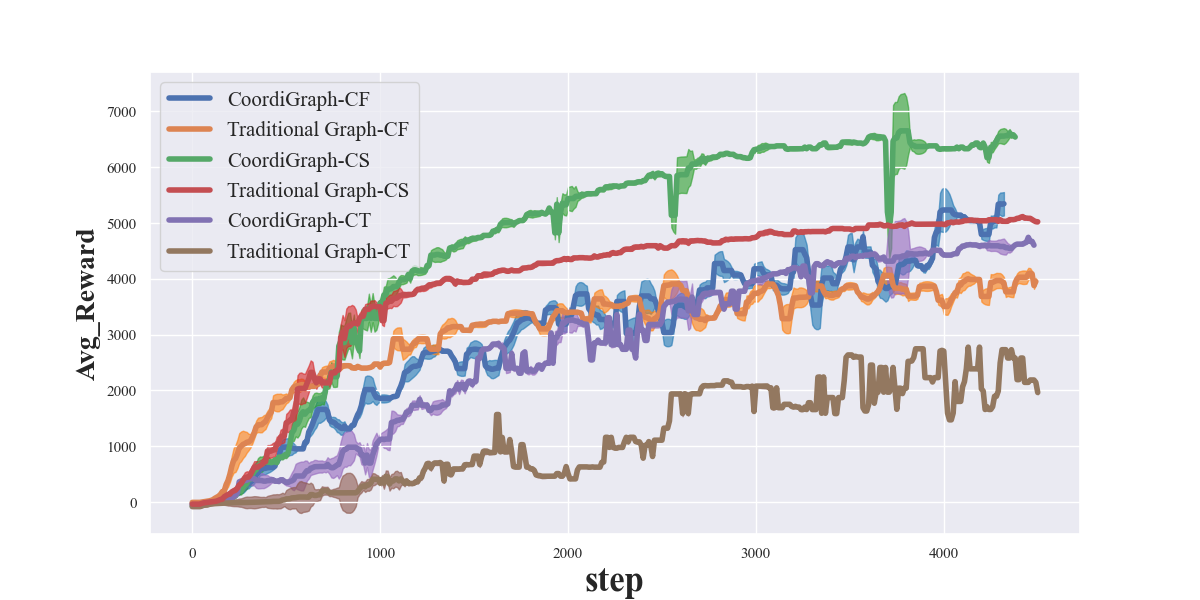}
        \vspace{-3ex}
    \caption{Results of comprehensive analysis on the complexity of agents}
    \label{Ab2}
    \vspace{-3ex}
\end{figure}

\textbf{Complexity of Agents: A Comprehensive Analysis}
In this experiment, we increased the structural complexity of agents by adding more connections in different parts. This aimed to assess CoordiGraph's generalization ability under varying levels of complexity. More connections provided additional parameters and richer feature representations, enabling the model to better understand the environment and learn complex strategies. We then tested these models with different complexities in the same environment.

By evaluating the performance and generalization ability in the same environment, we can understand how CoordiGraph performs under different levels of structural complexity in Figure \ref{Ab2}. Analyzing agent behavior and contrasting learning outcomes facilitates evaluating their generalization capabilities.

The results show that as agents become more complex, the model's ability to adapt remains stronger than the baseline. This suggests that CoordiGraph greatly improve complex models' ability to adapt to different environments and tasks. The model becomes better at understanding the environment and learning complex strategies. Increased complexity improves generalization, highlighting the advantage of CoordiGraph in helping complex models adapt to diverse environments and tasks.

\begin{figure}[h]
    \centering
    \begin{minipage}[t]{1.0\linewidth}
    \centering
        \begin{tabular}{@{\extracolsep{\fill}}c@{}c@{}c@{}@{\extracolsep{\fill}}}
        \includegraphics[width=0.40\linewidth,height=3.2cm]{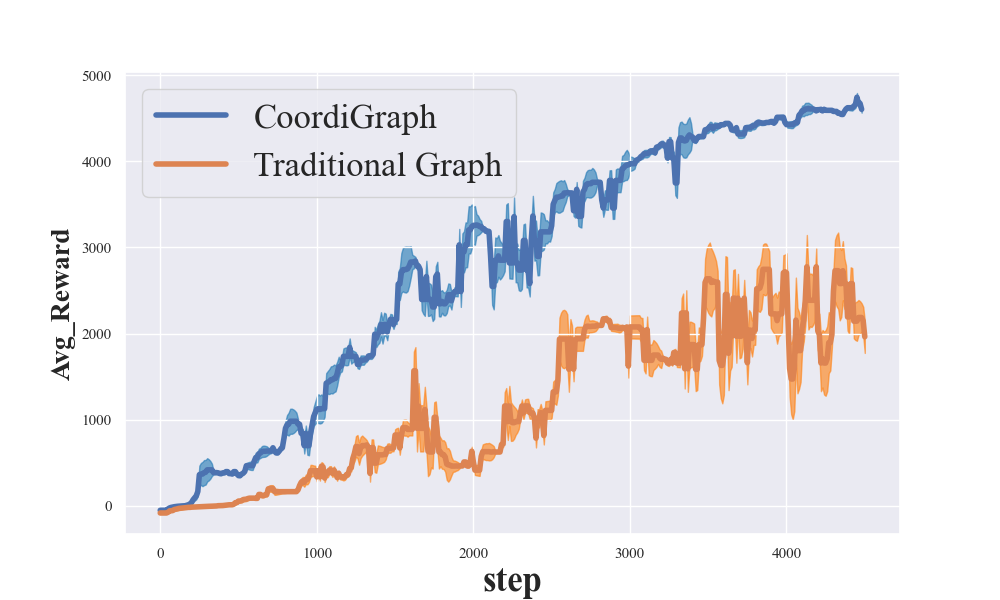} &
        \includegraphics[width=0.34\linewidth,height=3.2cm]{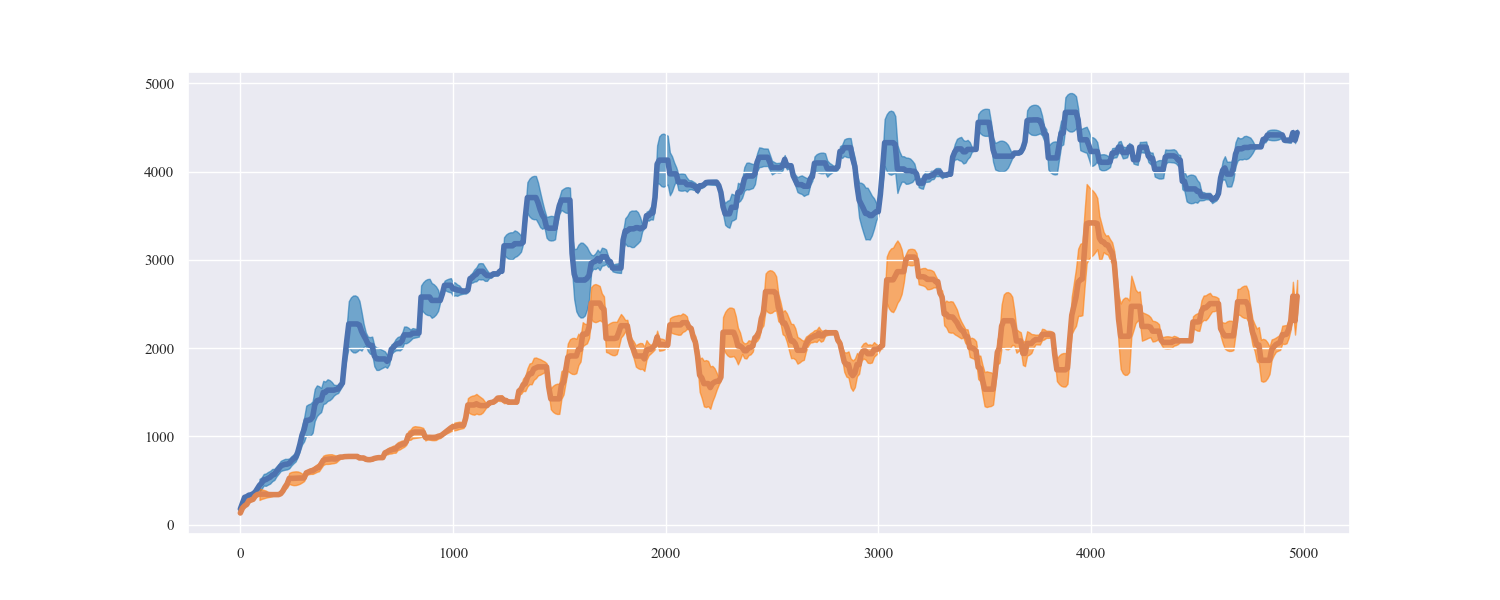}&
        \includegraphics[width=0.31\linewidth,height=3.2cm]{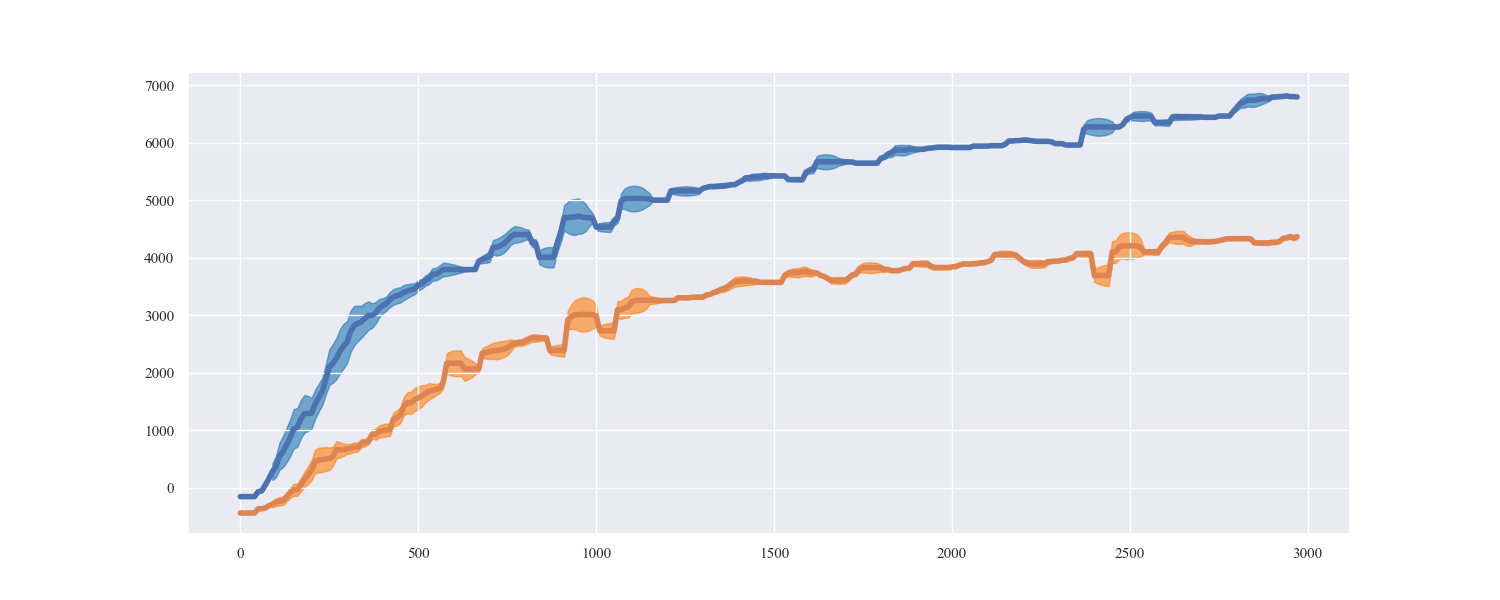}\\
            (a)CentipedeTen & (b)Walker2D & (c)HalfCheetah\\
        \end{tabular}
    \end{minipage}
    \caption{Results on the generalization of graph neural networks}
    \label{Ab3}
    \vspace{-1ex}
 \end{figure}

\textbf{Generalization of Graph Neural Networks}
We conducted experiments comparing CoordiGraph and traditional graph network to assess their performance in generalization and reward acquisition. We tested the models' ability to adapt to unfamiliar environments by introducing them to new tasks and scenarios that were different from the training data. This allowed us to evaluate how well the models could adapt and perform in diverse and novel conditions in Figure \ref{Ab3}.
\vspace{-2ex}


Based on our analysis, traditional graph network performs better in generalization, while CoordiGraph excels in both generalization and reward acquisition. CoordiGraph demonstrates superior adaptability and learning in unknown situations and new tasks, leading to better generalization. It also outperforms traditional graph network in reward acquisition by effectively understanding and utilizing reward signals to maximize rewards, resulting in faster learning and optimized agent coordination. In contrast, traditional graph network's performance in this area is relatively weaker, potentially leading to inaccurate action selection or suboptimal reward maximization.

\begin{figure}[h]
    \centering
    \begin{tabular}{@{\extracolsep{\fill}}c@{}c@{\extracolsep{\fill}}}
            \includegraphics[width=0.49\linewidth, height=3.6cm]{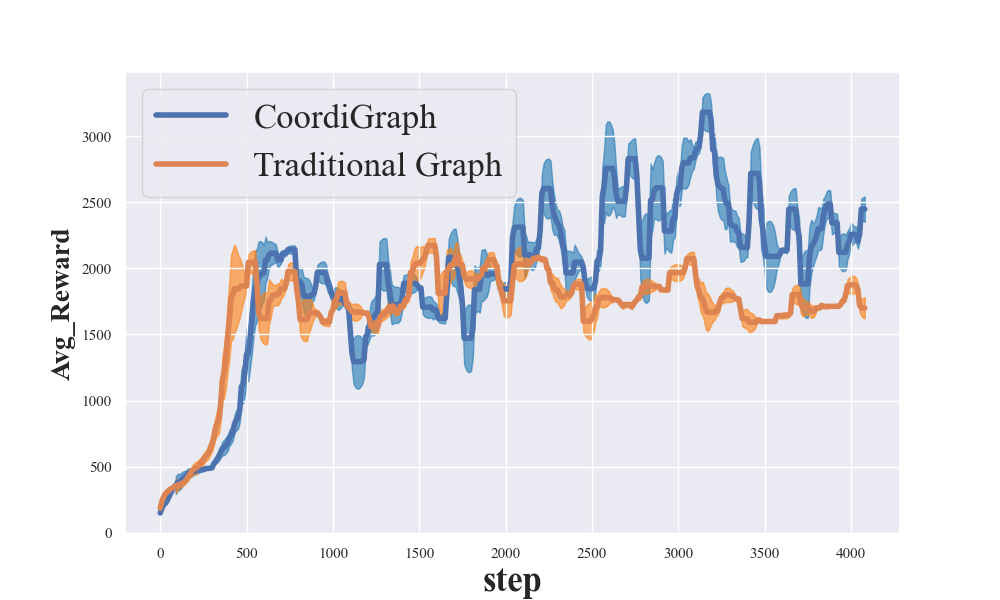} &
            \includegraphics[width=0.49\linewidth, height=3.6cm]{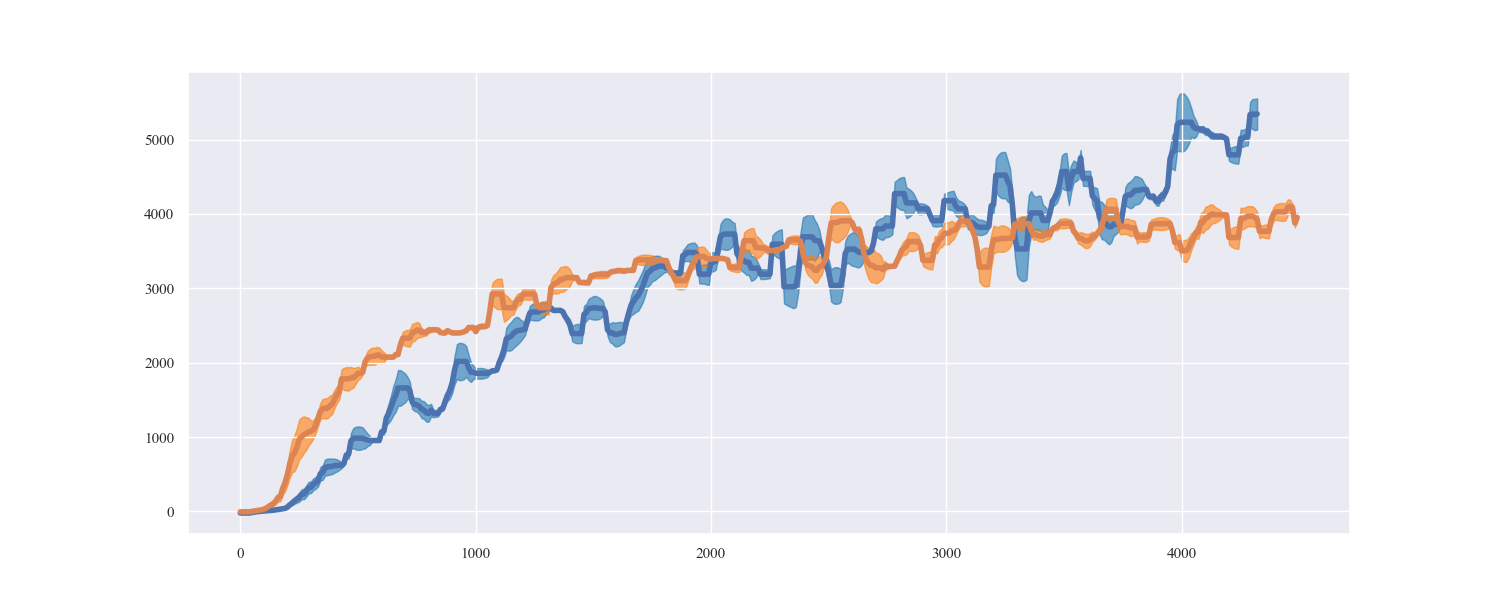}\\
            (a)Hopper & (b)CentipedeFour\\
    \end{tabular}
    \caption{Results on effects of subequivariant}
    \label{Ab4}
    \vspace{-2ex}
 \end{figure}

\textbf{Effects of Subequivariant}
We conducted experiments to investigate the impact of subequivariance on graph neural network models in reinforcement learning. We systematically removed subequivariant components to assess their effect on model performance. In motion coordination tasks, joint coordination behavior often exhibits symmetry and equivariance. Our goal was to understand how introducing subequivariance affects the model's ability to capture these characteristics and improve algorithmic model accuracy and performance in Figure \ref{Ab4}.
\vspace{-2.5ex}


The experimental results show that models with subequivariance outperform models without subequivariant components in terms of coordination accuracy. This emphasizes the importance of subequivariance in enhancing coordination task accuracy. Additionally, models with subequivariance demonstrate better generalization abilities. They can adapt and learn effectively even in unfamiliar situations and new coordination tasks, leading to improved generalization. Subequivariance helps the model capture joint relationships and coordination behaviors more effectively, resulting in more stable training and control processes.

\section{CONCLUSIONS}
In this paper, we introduce a novel framework, CoordiGraph, which leverages the subequivariant property to address the challenges of weak inter-joint coupling in high-dimensional motion control tasks using reinforcement learning. Experimental results indicate that CoordiGraph outperforms several baseline methods in complex motion control scenarios. These findings hint at the potential of subequivariance as a method to enhance coordination in intricate motion control tasks.

\section{Acknowledgement}
This work is supported by the National Natural Science Foundation of China (62102241) and Shanghai Municipal Natural Science Foundation (23ZR1425400).

\bibliographystyle{ieeetran}


\end{document}